\documentclass{article}

\usepackage[preprint]{neurips_2024}

\usepackage[utf8]{inputenc} 
\usepackage[T1]{fontenc}    

\usepackage{algorithm}
\usepackage{algpseudocode}
\usepackage{url}            
\usepackage{booktabs}       
\usepackage{amsfonts}       
\usepackage{nicefrac}       
\usepackage{microtype}      
\usepackage{xcolor}         
\usepackage{fancyhdr,graphicx,amsmath,amssymb}

\usepackage{tikz}
\usepackage{physics}
\usepackage{amsmath}
\usepackage{tikz}
\usepackage{mathdots}
\usepackage{yhmath}
\usepackage{cancel}
\usepackage{color}
\usepackage{siunitx}
\usepackage{array}
\usepackage{multirow}
\usepackage{amssymb}
\usepackage{gensymb}
\usepackage{tabularx}
\usepackage{extarrows}
\usepackage{booktabs}
\usetikzlibrary{fadings}
\usetikzlibrary{patterns}
\usetikzlibrary{shadows.blur}
\usetikzlibrary{shapes.geometric, arrows}

\tikzstyle{layer} = [rectangle, rounded corners, minimum width=3cm, minimum height=1cm, text centered, draw=black, fill=blue!20]
\tikzstyle{conv} = [layer, fill=green!20]
\tikzstyle{fc} = [layer, fill=orange!20]
\tikzstyle{output} = [layer, fill=red!20]
\tikzstyle{arrow} = [thick,->,>=stealth]

\usepackage{subfigure}
\usepackage{placeins}
\usepackage{hyperref}       
\title{Hierarchical Orchestra of Policies}

\author{%
  Thomas P Cannon \thanks{ \texttt{https://github.com/x4nnon}, \space \texttt{https://bath-rl-group.netlify.app/people/tom-cannon}} \\
  Department of Computer Science\\
  University of Bath\\
  UK, Bath, BA2 7AY  \\
  \texttt{tc2034@bath.ac.uk} \\
  \And
  Özgür Şimşek \\
  Department of Computer Science\\
  University of Bath\\
  UK, Bath, BA2 7AY  \\
  \texttt{os435@bath.ac.uk} \\
}

\begin{document}

\maketitle

\begin{abstract}
  Continual reinforcement learning poses a major challenge due to the tendency of agents to experience catastrophic forgetting when learning sequential tasks. In this paper, we introduce a modularity-based approach, called Hierarchical Orchestra of Policies (HOP), designed to mitigate catastrophic forgetting in lifelong reinforcement learning. HOP dynamically forms a hierarchy of policies based on a similarity metric between the current observations and previously encountered observations in successful tasks. Unlike other state-of-the-art methods, HOP does not require task labelling, allowing for robust adaptation in environments where boundaries between tasks are ambiguous. Our experiments, conducted across multiple tasks in a procedurally generated suite of environments, demonstrate that HOP significantly outperforms baseline methods in retaining knowledge across tasks and performs comparably to state-of-the-art transfer methods that require task labelling. Moreover, HOP achieves this without compromising performance when tasks remain constant, highlighting its versatility.
\end{abstract}

\section{Introduction}
\label{sec:intro}

Neural networks are typically trained on data drawn independently and identically from a static distribution. While this approach works well in many cases, it becomes challenging in environments that are continuously changing or when new environments are introduced. In dynamic settings, such as reinforcement learning, robotics, or dialogue systems, models must adapt to new information while preserving knowledge from previous tasks \citep{parisi2019continual}. However, neural networks often suffer from catastrophic forgetting, where learning new tasks leads to the rapid loss of previously acquired knowledge. The ability to learn new skills while maintaining existing knowledge is referred to as continual learning \citep{ring1994continual}.

To address the challenge of catastrophic forgetting, researchers have developed three primary categories of methods. The first category, \textit{regularization-based}, works by constraining updates to network parameters, thereby penalizing deviations from learned weight values that are critical for previous tasks. Notable examples of this approach include Elastic Weight Consolidation (EWC) and Synaptic Intelligence (SI) \citep{kirkpatrick2017overcoming, zenke2017continual}. The second category, \textit{replay-based}, mitigates forgetting by periodically rehearsing past experiences, either through actual data or synthetic generations, ensuring that the network continues to perform well on earlier tasks \citep{rolnick2019experience, shin2017continual}. The third category, \textit{modularity-based}, addresses the issue by structurally separating the network into modules, with each module dedicated to a specific task, thereby minimizing interference between tasks, prominent examples of this method are Progressive Neural Networks (PNN) by \citep{rusu2016progressive} and adaptive multi-column stacked sparse denoising autoencoder (AMC-SSDA) by \cite{agostinelli2013adaptive}. Finally there are some methods which use a combination of these, for example, \citet{schwarz2018progress} uses an active network and a knowledge-base network similar to the modularity-based methods, however it periodically compresses knowledge from the active into the knowledge network using EWC regularisation.

Our method, the Hierarchical Orchestra of Policies (HOP), is a modularity-based approach and is most similar to PNN. However, unlike PNN, HOP does not rely on a task identifier during training, which makes it more suitable for domain-incremental learning \citep{van2019three}. Additionally, HOP differs in that it combines network probability outputs directly through hierarchical weightings, rather than using latent connections between networks. Finally, we demonstrate HOP at a significantly larger scale — 18 hierarchical policy levels compared to only 3 in PNN. We show that HOP performs comparably to PNN, even when PNN is provided with task labels while HOP is not. Furthermore, HOP mitigates catastrophic forgetting across several Procgen environments \citep{cobbe2020leveraging}, achieving notable improvements over Proximal Policy Optimization (PPO), a reinforcement learning algorithm with inherent regularization properties \citep{schulman2017proximal}.

\section{Hierarchical Orchestra of Policies}
\label{sec:HOP}

\textbf{Hierarchical Orchestra of Policies (HOP)} is a modularity-based deep learning framework designed to mitigate catastrophic forgetting when learning new tasks. In this framework, a \textit{task} is defined as a specific Markov Decision Process (MDP), where distinct levels within a procedurally generated environment, or levels across different environments, are considered separate tasks \citep{puterman2014markov}. Although HOP is task-agnostic, all tasks are treated as episodic.

HOP relies on reinforcement learning algorithms that output stochastic policies, represented as $\pi(a \mid s)$ \citep{sutton2018reinforcement}. In our work, PPO serves as the base algorithm for HOP. The framework introduces three key mechanisms to form and use a collection of policies: 

\begin{enumerate}
    \item \textit{Checkpoints} to freeze and store policies at a certain stage of training.
    \item \textit{Orchestration} of policy activation based on state similarity.
    \item \textit{Hierarchical weightings} to balance the influence of previous and new policies.
\end{enumerate}

These mechanisms enable the agent to recover and maintain performance across diverse tasks without significant interference, thereby promoting continual learning in complex environments.

\textbf{Checkpoints.} The agent initially learns a policy using a base algorithm. After $\mathcal{T}_{checkpoint}$ time-steps, HOP initializes a \textit{checkpoint}, where the current learning policy $\pi$ is frozen and evaluated in the currently available tasks. During checkpoint evaluation, if the episodic return $R$ surpasses a predefined threshold $R_{threshold}$, all states encountered during that task episode are stored in a set of \textit{trusted states} $S_m$ which is linked with the policy checkpoint $\pi_m$, where $m$ is the count of the checkpoint.

\textbf{The Orchestra.} When the agent resumes learning, it dynamically activates checkpoint policies $\pi_m$ determined by the similarity between the current state $s_t$ and any $s_m \in S_m$. If the current state $s_t$ is \textit{similar} to any state in $S_m$, then the corresponding frozen policy $\pi_m$ is activated ($I_{m} = 1$). Similarity is determined by a threshold value $\omega$, which, in all of our experiments, has been defined as any $s_m \in S_m$ with a cosine similarity greater than 0.98 with $s_t$.

Rather than selecting actions directly from the distribution of a single frozen policy ($a_t \sim \pi_m(s_t)$), which could lead to conflicts when multiple policy checkpoints are activated, HOP combines the distributions from activated policy checkpoints ($I_m\pi_m$) and the current learning policy $\pi_n$ into a joined action policy, denoted as $\pi_{n_a}$ (see equation \ref{eq:HOP}). Here, $n$ represents the current count of all policies, and the subscript $a$ denotes the combined policy from which the action is sampled. This approach allows the agent to leverage past knowledge while adapting to new tasks, promoting continual learning.

To avoid significant and undesired output shifts caused by small changes in the state, frozen policies predict actions based on the most similar state, $s_{m} \in S_m$, to the current state $s_t$ \citep{szegedy2013intriguing}. This state is referred to as $s^*_{m}$, and actions are chosen as $\pi_m(a \mid s^*_{m})$ rather than directly from $s_t$. This dynamic activation of multiple policies is called the \textit{orchestra} of policies, a term borrowed from \cite{jonckheere2023symphony} but applied differently in this work.

\begin{figure}[htbp]
    \centering
    \includegraphics[width=0.8\textwidth]{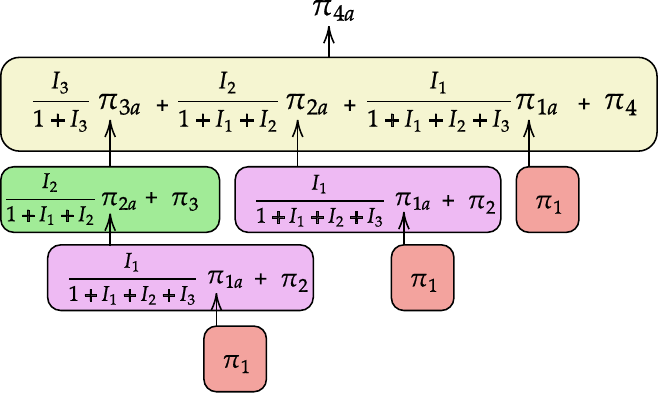}
    \caption{Hierarchical formation of the fourth level of a HOP action policy.}
    \label{fig:hierarchy_layout}
\end{figure}

\textbf{Hierarchical Weightings.} As the agent learns, it is expected to achieve higher task-specific rewards, which suggests that newer policies for the same tasks are likely to outperform older policy checkpoints for the same task. Thus, simply averaging all policies, as represented by $\pi_{n_a} = \frac{1}{n}\sum_{m=0}^{n} \pi_{m}$, is impractical. Moreover, because the agent does not know the identities of tasks, multiple policy checkpoints may activate; therefore, simply sampling actions from newer policy checkpoints is not possible. To address this, HOP introduces a hierarchical discount factor, denoted as $W$, to determine the contributions of policy checkpoints. Each joined action policy at time of checkpoint is a combination of previous policy checkpoints, creating a hierarchical structure, as shown in Figure \ref{fig:hierarchy_layout} and Equation \ref{eq:HOP}. This hierarchical structure assigns a higher weight to more recent activated policies. For some examples of how this concept functions, please refer to Appendix \ref{app:hw_examples}.

\textbf{Policy Updates}: The update process for the learning policy ($\pi_n$) follows the same procedure as that used by the base algorithm. Specifically, all necessary attributes associated with the policy are sampled from $\pi_n$, except for the action, which is instead sampled from the current joined action policy ($\pi_{n_a}$). We also allow gradients to propagate to the policy checkpoints $\pi_m$, but only for the states where they are activated. For more detail and pseudo code refer to Appendix \ref{app:algorith_details}.

The HOP action policy is expressed as,

\begin{align}
\mathcal{\pi}_{n_a}(s_t) &= \mathcal{\pi}_n(s_t) + \sum_{m=1}^{n-1} W_m \pi_{m_a}(s^*_{m}),\quad \label{eq:HOP}\\
W_m &= \frac{I_m}{1+\sum_{k=m}^{M} I_k}, \\[10pt]
I_{m} &= 
\begin{cases} 
1 & \text{if }   \frac{s^*_{m} \cdot s_t}{\|s^*_{m}\| \|s_t\|} > \omega. \\
0 & \text{otherwise}.
\end{cases}
\end{align}

Here, $s^*_{m}$ is the most similar state from all $s \in S_m$, and $\pi_{m_a}$ represents the logits of the $m$-th joined action policy checkpoint. $M$ is the total number of checkpoints conducted. $n$ is the current count of all policies. $I_m$ is the activation related to each policy $\pi_m$. And $\omega$ is the similarity threshold of the current state with previous states in $S_m$.

\begin{figure}[htbp]
    \centering
    \includegraphics[width=0.8\textwidth]{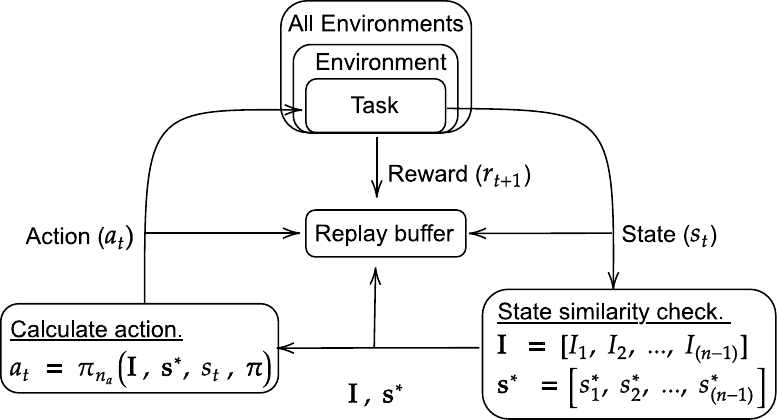}
    \caption{The flow of information as a HOP agent acts in a task.}
    \label{fig:hop_diagram}
\end{figure}

Figure \ref{fig:hop_diagram} depicts the flow of information within the HOP framework, illustrating how the agent evaluates observations, selects relevant policies, and takes actions to adapt continually across tasks. The agent begins by assessing the current state \( s_t \) against a similarity threshold for each checkpoint policy \( \pi_m \in \boldsymbol{\pi} \), where \( \boldsymbol{\pi} \) denotes all previously stored policies. For each policy, it identifies the most similar reference state \( s^*_m \in S_m \) and calculates an activation \( I_{m} \) based on this similarity. If the similarity metric—calculated as the cosine similarity between \( s_t \) and \( s^*_m \)—exceeds the predefined threshold \( \omega \), the corresponding policy \( \pi_m \) is activated (\( I_{m} = 1 \)), otherwise, it remains inactive (\( I_{m} = 0 \)). The activated policies then contribute to the current joint action policy \( \pi_{n_a} \), which combines outputs from the activated checkpoint policies \( I_m\pi_m \) and the current learning policy \( \pi_n \), weighted hierarchically according to their relative recency and activations, as shown in Equation \ref{eq:HOP}. 

The agent samples actions from \( \pi_{n_a} \) rather than directly from the learning policy \( \pi_n \), enabling it to leverage past knowledge while adapting to new tasks. The chosen action interacts with the environment, producing a new state and a corresponding reward. These states, actions, rewards, and activations are stored in the replay buffer to support continual learning and mitigate forgetting by allowing the agent to revisit past experiences. During training, the replay buffer’s stored activations also guide the gradient computations, allowing only the activated policies to contribute to the policy update. This targeted update process refines gradient flow selectively based on activation, promoting modularity and stability in learning across diverse task environments.

\section{Results}
\label{sec:Results}

We evaluated the performance of HOP using the Procgen suite of environments \citep{cobbe2020leveraging}. The experimental setup consisted of three phases of training. In the first phase, the agent trains for three million time-steps on multiple levels of a selected Procgen environment to develop its ability to learn and generalize. In the second phase, the environment was switched to a different one, and the agent continued training for another three million time-steps, assessing its adaptability and ability to transfer learning. Finally, in the third phase, the agent returned to the original environment for an additional three million time-steps to evaluate retention of skills and re-adaptation. Throughout training, the agent's objective is to optimize the reward functions defined by the Procgen environments, which typically involve maximizing cumulative rewards for task-specific objectives such as reaching goals, collecting items, or avoiding obstacles. This is a simplified experimental set-up to that conducted by \citet{schwarz2018progress} in their examination of P\&C. 

We conducted experiments with three different environment combinations: StarPilot and Climber, Ninja and StarPilot, and Ninja and CoinRun, repeating each with four random seeds. During training, periodic evaluation episodes were performed to measure performance, and \textit{checkpoints} were saved every 500,000 time-steps.

\begin{figure}[h]
    \centering

    \includegraphics[width=0.3\textwidth]{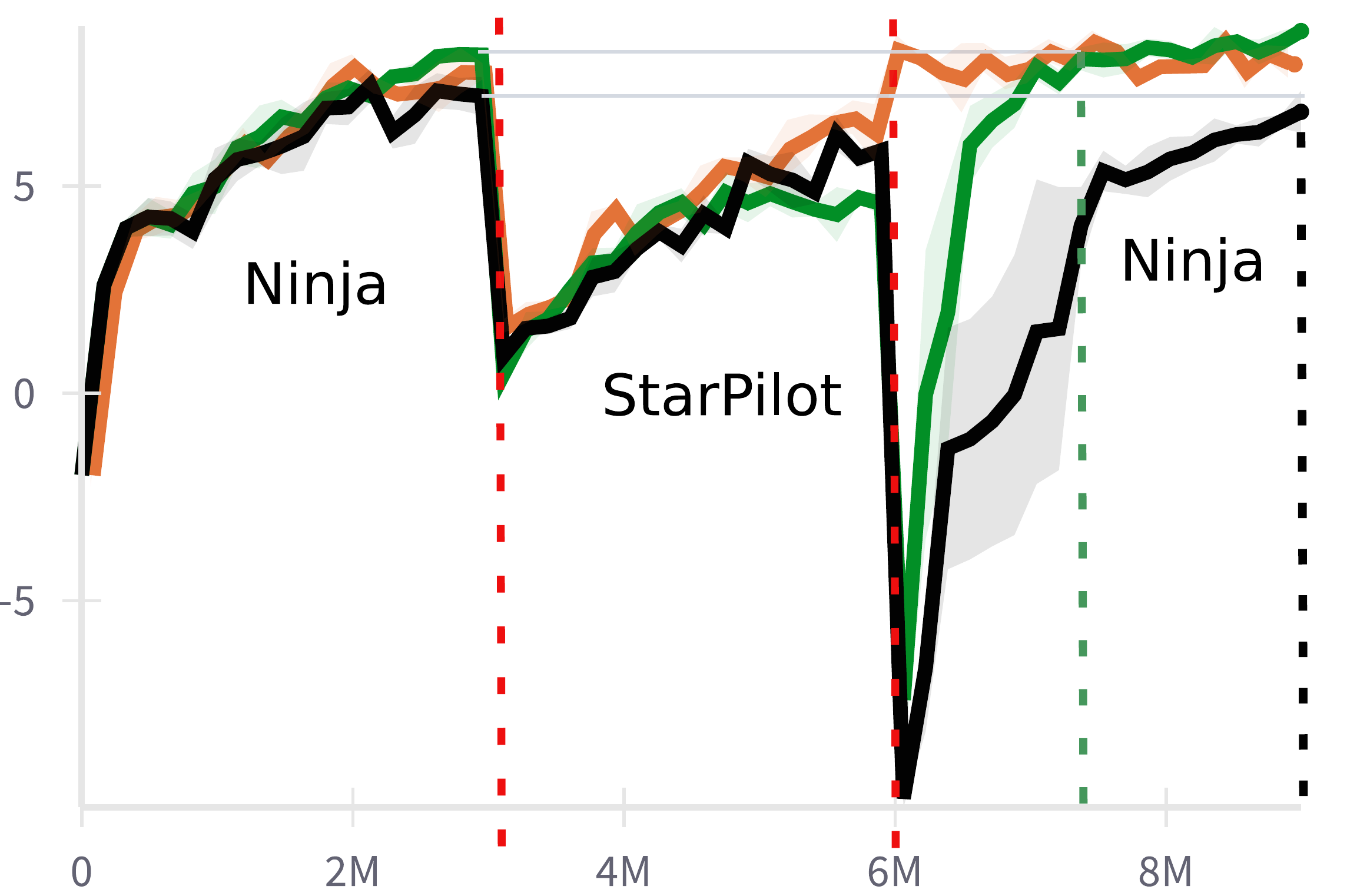}
    \hfill 
    \includegraphics[width=0.35\textwidth]{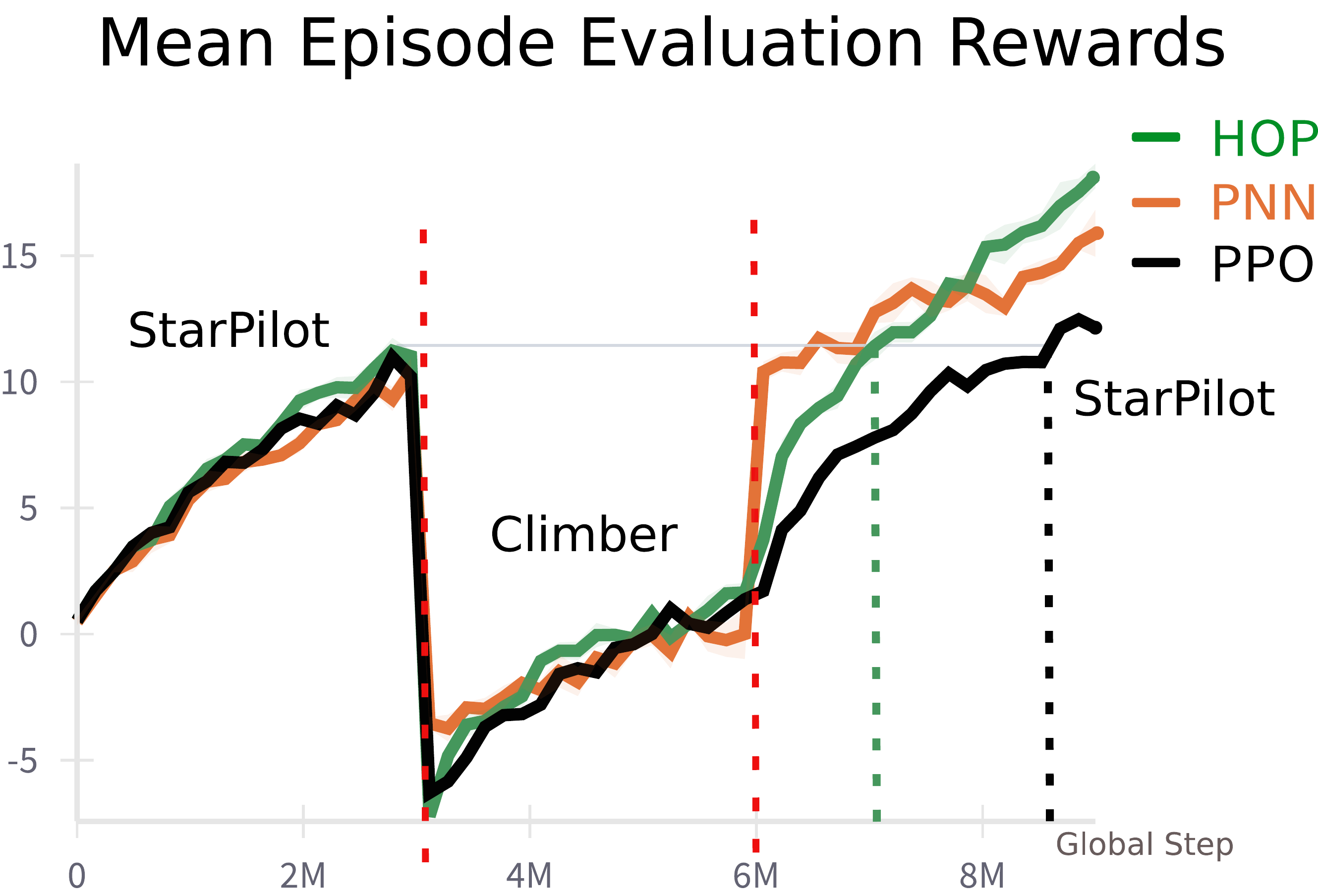}
    \hfill
    \includegraphics[width=0.3\textwidth]{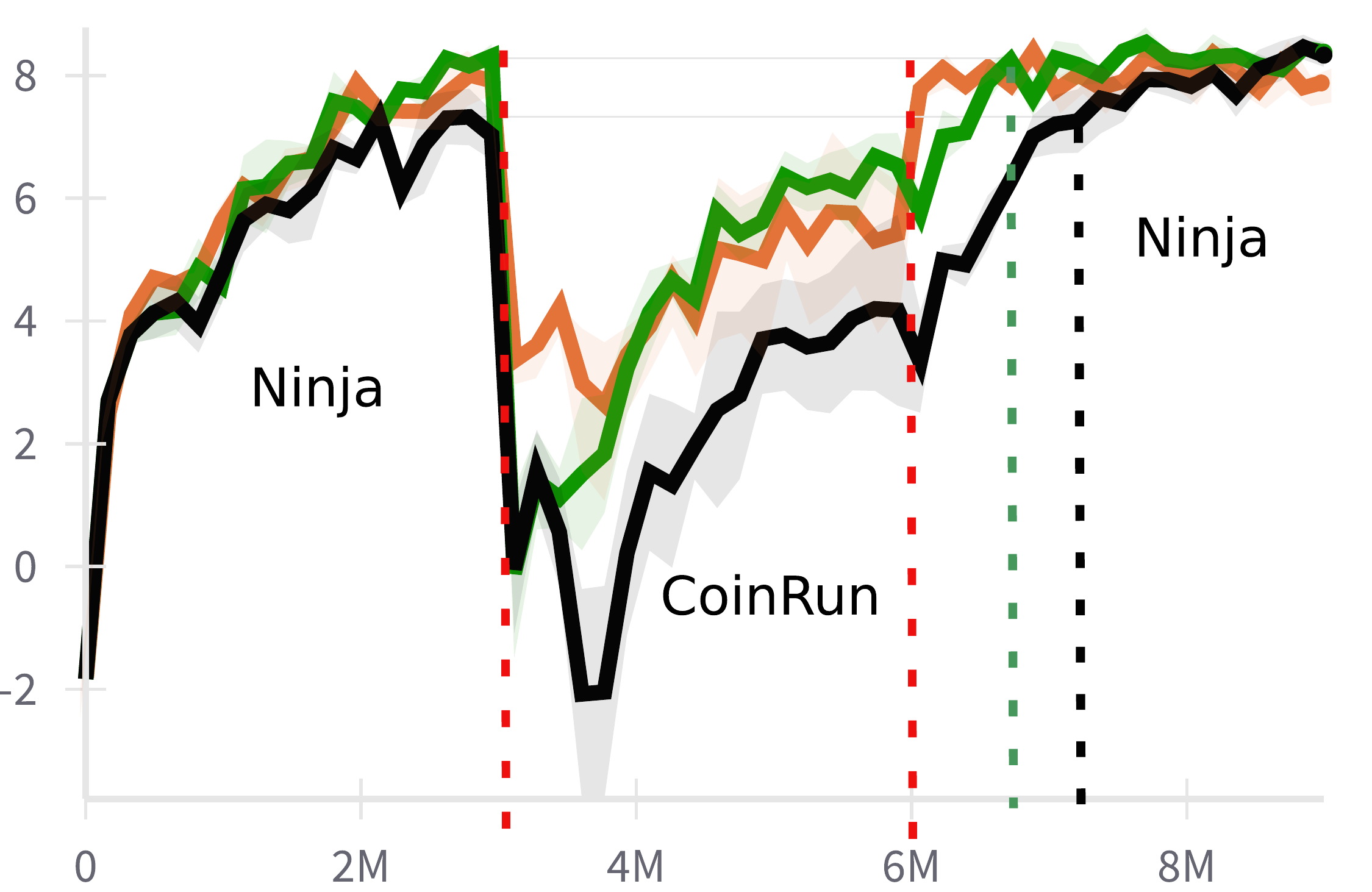}
    
    \caption{Training performance of HOP, PNN and PPO on three experiments where environments are periodically changed. The red dashed lines indicate the points when the environment are switched. The green dashed lines show when HOP returns to the highest average evaluation reward achieved in the first environment before the change. The black dashed lines represents this point for PPO. Shaded areas are the standard error. All experiments are conducted with the Procgen easy setting.}
    \label{fig:results_change}
\end{figure}

HOP was compared with standard PPO and a modified version of Progressive Neural Networks (PNN) for use with PPO -- see appendix \ref{app:PNN_details} for full details of the modifications. We allowed PNN to have task identifiers but not HOP. Results presented in Figure~\ref{fig:results_change} indicate that HOP outperformed PPO in both the rate of performance recovery and the final averaged evaluation return after training. We found that HOP had comparable performance to PNN in all but the very beginning of the third phase of learning. Table~\ref{tab:comparison} summarizes the total steps after the second phase of learning required for each method to recover to the performance level achieved at the end of the initial phase of training, and the final averaged evaluation return. For a complete description of the experiments and environments please see appendix \ref{app:experiment_details}.

\begin{table}[h]
    \centering
    \begin{tabular}{lccccc}
        \toprule
        & \multicolumn{2}{c}{\textbf{Steps-to-return ($10^6$)}} & \multicolumn{3}{c}{\textbf{Final Rewards}} \\
        \cmidrule(lr){2-3} \cmidrule(lr){4-6}
        & PPO & HOP & PPO & HOP & PNN \\
        \midrule
        StarPilot - Climber & 2.68 & \textbf{1.04} \tiny{-61.2\%} & 12.14 & \textbf{18.15} \tiny{49.5\%} & 15.98 \tiny{31.6\%}\\
        Ninja - StarPilot & 3+ & \textbf{1.70} \tiny{-43+\%} & 6.79 & \textbf{8.73 }\tiny{28.6\%} & \textbf{7.97} \tiny{17.37\%}\\
        Ninja - Coinrun & 1.37 & \textbf{0.72} \tiny{-47.7\%} & 8.33 & 8.37 \tiny{0.48\%} & 7.83 \tiny{-6.00\%}\\
        \bottomrule
    \end{tabular}
    \vspace{0.1cm}
    \caption{A comparison of PPO, HOP, and PNN. Steps-to-return represents the number of steps (in millions) to re-acquire the same average evaluation reward at the end of the first period of learning in that environment, PNN is not included in these comparisons as it uses a separate actor and critic network per task. Final rewards display the final average evaluation rewards at the end of all training. The percentages show the difference compared to the baseline PPO method.}
    \label{tab:comparison}
\end{table}

\section{Summary and Discussion}
\label{sec:Discussion}

We present a novel modularity-based approach, the Hierarchical Orchestra of Policies, to address catastrophic forgetting in continual life-long reinforcement learning. In our empirical evaluation, HOP outperforms PPO in continual learning scenarios, achieving a faster recovery of performance and final performance. Both HOP and PNN demonstrate substantial transfer between environments with similar dynamics and state spaces such as Ninja and CoinRun. In these scenarios HOP can activate relevant frozen policies learned from Ninja while acting in CoinRun, similar to PNN’s \textit{adapter} networks connecting separate \textit{columns}. However, unlike PNN, HOP does not require task labels, making it more versatile for real-world applications where task boundaries are not clearly defined.

However, the effectiveness of HOP depends on the careful tuning of some hyper-parameters, particularly the similarity threshold ($w$) and reward threshold ($P$), which must be set appropriately for all expected tasks. See Appendix \ref{app:experiment_details} 

Future work could expand HOP’s evaluation by testing transitions between highly diverse tasks and environments where task boundaries are ambiguous, a setting in which PNN and similar methods are less effective. Additionally, HOP could be adapted to continuous environments with fluid task transitions, further highlighting its robustness in real-world scenarios. To address performance drops immediately following task distribution changes, a learnable parameter could be introduced which could dynamically adjust the influence of previous checkpoints, enabling immediate adaptation while maintaining learning.

\newpage
\bibliographystyle{apalike}
\bibliography{bib}

\newpage 
\appendix

\section{Appendix}

\subsection{HOP Algorithm details}
\label{app:algorith_details}

The logic provided in algorithm \ref{alg:HOP}, is suitable for use with PPO (or any other actor critic style base function. It is expected that HOP would work with any method with a stochastic policy, however this has yet to be tested. Table \ref{tab:hop_parameters} details all of the extra parameters that HOP requires.

\newcommand{\boxcomment}[1]{\hfill \textcolor{gray}{\fbox{\footnotesize #1}}}
\newcommand{\boxcommentblue}[1]{\hfill \textcolor{blue}{\fbox{\footnotesize #1}}}

\begin{algorithm}[h]
    \caption{Hierarchical Orchestra of Policies (HOP) with PPO}
    \label{alg:HOP}
    
    \textbf{Initialize:} Current hierarchy depth $n$ = 1, Policy $\pi_n$ == $\pi_{n_a}$, similarity threshold $w$, and reward threshold $P$, total steps $D$, step = 0, checkpoint interval $C$, state $s_t$, done = 0, value function $V_{\theta}$, batch size $T$, buffers $B$, and other PPO parameters $\phi$.
    
    \begin{algorithmic}[1]
        \State \textbf{Training:}
        \While{step < $D$}
            \While{step < $T$}
                \For{each frozen policy $\pi_m$} \boxcommentblue{Activation logic}
                    \If{cosine similarity $\frac{s_{m_{\text{max-sim}}} \cdot s_t}{\|s_{m_{\text{max-sim}}}\| \|s_t\|} > w$}
                        \State Activate policy $\pi_m$, set $A_m = 1$
                    \Else
                        \State Deactivate policy $\pi_m$, set $A_m = 0$
                    \EndIf
                \EndFor
                \State Sample action $a_t \sim \pi_{n_a}(s_t)$ as per equation \ref{eq:HOP}
                \State $s_t$, reward, done = environment.step
                \State $B \gets s_t$, reward, done, $a_t$ 
                \If{done}
                    \State reset environment
                \EndIf
            \EndWhile
            \State Update $\pi_n$ and $V_{\theta}$ from $B$ using PPO algorithm \boxcomment{\tiny{Allow gradients to propagate to activated policies in those states only.}}
            \If{step == C} \boxcommentblue{Checkpoint logic}
                \State Freeze policy $\pi_n$ as $\pi_m$
                \State Evaluate $\pi_{m_a}$ on all currently available tasks 
                
                \boxcomment{\tiny{Evaluation can be from the most recent experiences or from running new evaluations. We use new evaluations in our experiments}}
                \For{each evaluation episode}
                    \If{episodic return $R > P$}
                        \State Append all states in evaluation to $S_m$
                    \EndIf
                \EndFor
            \EndIf
        \EndWhile
    \end{algorithmic}
\end{algorithm}

\begin{table}[ht]
\centering
\begin{tabular}{|c|l|c|l|}
\hline
\textbf{Symbol} & \textbf{Name}              & \textbf{Default} & \textbf{Notes}                                                 \\ \hline
$P$               & Reward threshold           & 7.5              & \small{Can be generalized if rewards are normalized}     \\ \hline
$w$               & Similarity threshold       & 0.98               & \small{Cosine similarity}                                                               \\ \hline
$C$               & Checkpoint interval        & 500,000              & \small{Time-steps. In our experiments forms 18 policies}                                                               \\ \hline
\end{tabular}
\caption{All additional parameters for the HOP algorithm.}
\label{tab:hop_parameters}
\end{table}

\FloatBarrier

\subsection{Experiment details}
\label{app:experiment_details}

\begin{figure}[h]
    \centering

    \includegraphics[width=0.9\textwidth]{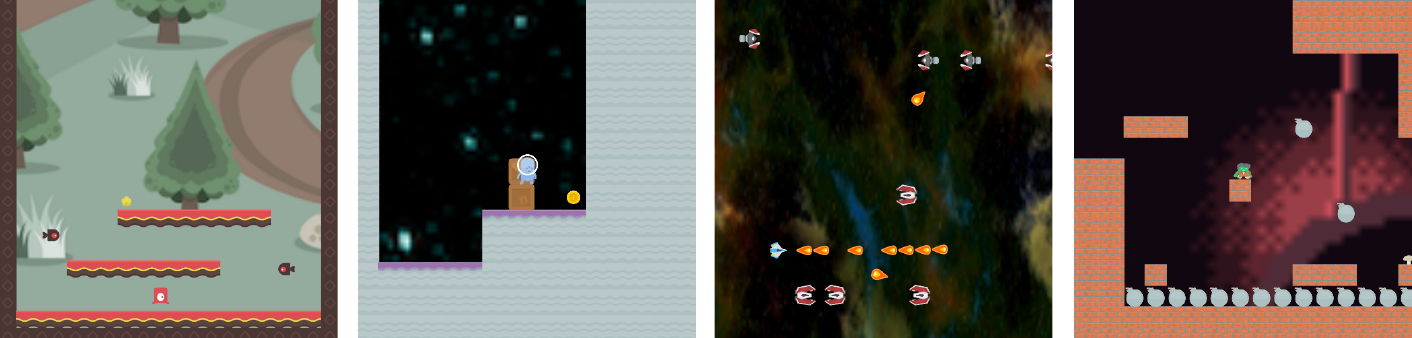}

    \caption{From left to right, Climber, CoinRun, StarPilot and, Ninja. In our experiments the backgrounds are all black (use\_backgrounds=False).}
    \label{fig:procgen_examples}
\end{figure}

Our experiments are conducted in the Procgen suite of environments introduced by \citet{cobbe2020leveraging}. Specifically, we use Ninja, StarPilot, Climber, and CoinRun as our environments. These can be viewed at \href{https://github.com/openai/procgen}{https://github.com/openai/procgen}, and we also provide a table with a snapshot of each environment in Figure \ref{fig:procgen_examples}. In Procgen, there are options that can reduce the complexity of the environments. We activate the following options: \texttt{use\_backgrounds=False}, \texttt{restrict\_themes=True}, \texttt{distribution\_mode=easy}, and \texttt{use\_sequential\_levels=True}. However, we do not activate \texttt{use\_monochrome\_assets}, as we found that it lacked proper indications for agent direction. In all of our experiments, the agent's goal was to maximize the cumulative reward provided by the environment. The state is represented as an 84x84 pixel image, and the agent has 15 possible actions.

We run the same experiment with different combinations of environments. The experiment is conducted in three phases, each evenly distributed over the total number of time steps ($X$): 
\begin{enumerate}
    \item The agent trains in environment 1 with $T_1$ tasks in distribution.
    \item Learning is switched to environment 2 with $T_2$ tasks.
    \item Learning is switched back to environment 1 for the same $T_1$ tasks.
\end{enumerate}

In our experiments, $X = 9,000,000$, and $T_1 = T_2 = 30$. PNN is given a task identifier for each environment, enabling it to use the correct networks and adapters. HOP, on the other hand, does not require these and is not given them. Every 163,840 time steps, the agent is evaluated in the current distribution of tasks, which in this case consists of 30 Procgen levels in the current training environment of that phase, the reward in this evaluation phase is the cumulative reward - 0.01*total steps taken in the environment, which gives a better indication of efficiency. The results we report are based on these evaluated tasks. Conducting evaluation episodes at fixed intervals provides the clearest and most accurate representation of agent performance.

The three experiments shown in Figure \ref{fig:results_change} are conducted using the combinations of environments listed in Table \ref{tab:experiment_phases}. In the first experiment, the environments are completely different, with distinct dynamics and little to no shared understanding between them. For the second experiment, we believed there would be some overlap; while StarPilot scrolls from right to left, Climber scrolls vertically from bottom to top. The final experiment features considerable shared dynamics, as both Ninja and CoinRun are platforming games. We hypothesized that this setup would demonstrate the transfer and recovery of performance across different levels of difficulty. However, we observed that only the Ninja and CoinRun experiment exhibited meaningful transfer for both PNN and HOP.

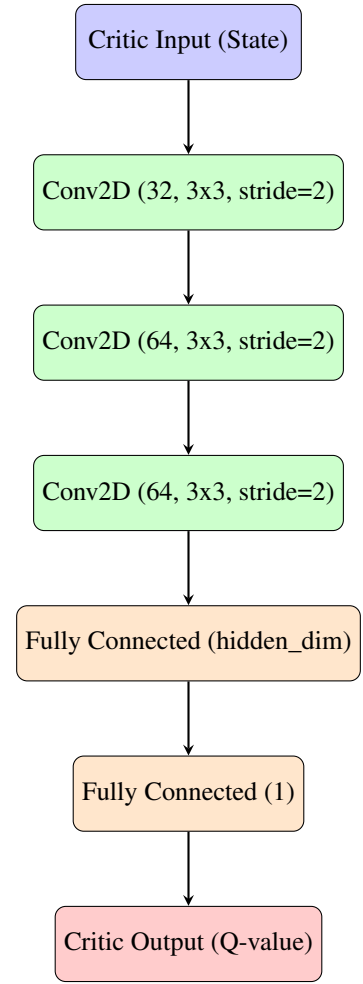
\begin{figure}[h!]
    \centering
    \begin{tikzpicture}[node distance=2cm]

    \node (inputA) [layer] {Actor Input (State)};
    \node (conv1A) [conv, below of=inputA] {Conv2D (32, 3x3, stride=2)};
    \node (conv2A) [conv, below of=conv1A] {Conv2D (64, 3x3, stride=2)};
    \node (conv3A) [conv, below of=conv2A] {Conv2D (64, 3x3, stride=2)};
    \node (fc1A) [fc, below of=conv3A] {Fully Connected (hidden\_dim)};
    \node (fc2A) [fc, below of=fc1A] {Fully Connected (num\_actions)};
    \node (outputA) [output, below of=fc2A] {Actor Output (Action)};

    \draw [arrow] (inputA) -- (conv1A);
    \draw [arrow] (conv1A) -- (conv2A);
    \draw [arrow] (conv2A) -- (conv3A);
    \draw [arrow] (conv3A) -- (fc1A);
    \draw [arrow] (fc1A) -- (fc2A);
    \draw [arrow] (fc2A) -- (outputA);

    \node (inputC) [layer, right of=inputA, xshift=6cm] {Critic Input (State)};
    \node (conv1C) [conv, below of=inputC] {Conv2D (32, 3x3, stride=2)};
    \node (conv2C) [conv, below of=conv1C] {Conv2D (64, 3x3, stride=2)};
    \node (conv3C) [conv, below of=conv2C] {Conv2D (64, 3x3, stride=2)};
    \node (fc1C) [fc, below of=conv3C] {Fully Connected (hidden\_dim)};
    \node (fc2C) [fc, below of=fc1C] {Fully Connected (1)};
    \node (outputC) [output, below of=fc2C] {Critic Output (Q-value)};

    \draw [arrow] (inputC) -- (conv1C);
    \draw [arrow] (conv1C) -- (conv2C);
    \draw [arrow] (conv2C) -- (conv3C);
    \draw [arrow] (conv3C) -- (fc1C);
    \draw [arrow] (fc1C) -- (fc2C);
    \draw [arrow] (fc2C) -- (outputC);

    \node at (-2.5,1) {\textbf{ProcGen Actor}};
    \node at (6.5,1) {\textbf{ProcGen Critic}};

    \end{tikzpicture}
    \caption{Separate Actor and Critic Networks for the ProcGen Architecture}
    \label{fig:actor_critic_architecture}
\end{figure}

We use PPO as a baseline algorithm and as the foundation for both HOP and PNN. Our PPO implementation is based on the version by \citet{huang2022cleanrl}. The only modification we made was separating the actor and critic networks, which we found easier to work with and which outperformed the shared convolutional layer approach. Figure \ref{fig:actor_critic_architecture} illustrates our implementation. We kept the PPO-specific hyperparameters fixed for its use with HOP, PNN, and the base PPO. These hyperparameters were optimized for the base PPO, and while a small benefit might have been observed for HOP and PNN if a hyperparameter sweep had been conducted, both performed as expected, so we did not pursue this. The PPO-specific hyperparameters are shown in Table \ref{tab:ppo_hyperparameters}, and all other relevant parameters are shown in Table \ref{tab:other_parameters}.

\begin{table}[ht]
\centering
\begin{tabular}{|c|c|c|c|}
\hline
\textbf{Experiment Number} & \textbf{Phase 1} & \textbf{Phase 2} & \textbf{Phase 3} \\ \hline
1                          & Ninja            & StarPilot        & Ninja            \\ \hline
2                          & StarPilot        & Climber          & StarPilot        \\ \hline
3                          & Ninja            & CoinRun          & Ninja            \\ \hline
\end{tabular}
\caption{Experiment Phases for Different Environments}
\label{tab:experiment_phases}
\end{table}

\begin{table}[ht]
\centering
\begin{tabular}{|l|l|l|}
\hline
\textbf{PPO Hyperparameters} & \textbf{Description} & \textbf{Value} \\ \hline
gamma                        & Discount factor for future rewards         & 0.999          \\ \hline
vf\_coef                     & Weight of value function loss              & 0.5            \\ \hline
ent\_coef                    & Weight of entropy bonus                    & 0.01           \\ \hline
norm\_adv                    & Normalize advantages during optimization   & true           \\ \hline
num\_steps                   & Number of steps per environment per update & 256            \\ \hline
clip\_coef                   & Clipping factor for policy loss            & 0.2            \\ \hline
gae\_lambda                  & Generalized Advantage Estimation parameter & 0.95           \\ \hline
batch\_size                  & Total number of samples per batch          & 16384          \\ \hline
clip\_vloss                  & Clip value function loss                   & false          \\ \hline
target\_kl                   & Target KL divergence                       & 0.05           \\ \hline
update\_epochs               & Number of optimization epochs per update   & 3              \\ \hline
minibatch\_size              & Size of mini-batches used in optimization  & 2048           \\ \hline
max\_grad\_norm              & Maximum gradient norm for clipping         & 0.5            \\ \hline
anneal\_lr                   & Whether to anneal learning rate over time  & false          \\ \hline
num\_envs                    & Number of parallel environments            & 64             \\ \hline
num\_minibatches             & Number of mini-batches per optimization step & 8            \\ \hline
\end{tabular}
\caption{PPO Hyperparameters}
\label{tab:ppo_hyperparameters}
\end{table}

\begin{table}[ht]
\centering
\begin{tabular}{|l|l|l|}
\hline
\textbf{Other Parameters}     & \textbf{Description}                            & \textbf{Value} \\ \hline
cuda                         & Use CUDA for computation                        & true           \\ \hline
easy                         & Procgen easy parameter                          & 1              \\ \hline
proc\_start                  & Starting level in Procgen                       & 1              \\ \hline
reward\_limit                & Checkpoint minimum reward limit ($P$)           & 7.5             \\ \hline
report\_epoch                & Number of steps between evaluation reports    & 163840         \\ \hline
learning\_rate               & Learning rate for optimizer                     & 0.0005         \\ \hline
max\_ep\_length              & Maximum number of steps per episode             & 1000           \\ \hline
use\_monochrome              & Whether to use monochrome assets in environment & 0              \\ \hline
eval\_batch\_size            & Number of episodes for evaluation               & 30             \\ \hline
max\_eval\_ep\_len           & Maximum length of episodes during evaluation    & 1000           \\ \hline
proc\_num\_levels            & Number of levels in the Procgen environment     & 30             \\ \hline
total\_timesteps             & Total number of timesteps for training          & 9000000        \\ \hline
eval\_specific\_envs         & Number of environments used for evaluation      & 30             \\ \hline
torch\_deterministic         & Enable deterministic operations in PyTorch      & true           \\ \hline
min\_similarity\_score       & Minimum cosine similarity for activation $w$        & 0.98           \\ \hline
checkpoint\_interval          & Minimum cosine similarity for activation $C$        & 500,000         \\ \hline
\end{tabular}
\caption{Other Experiment Parameters}
\label{tab:other_parameters}
\end{table}

\FloatBarrier

\subsection{PNN with PPO Algorithm Details}
\label{app:PNN_details}

Progressive Neural Networks (PNN) were introduced by \citet{rusu2016progressive}. In their paper, they describe how separate policy networks (referred to as columns) and links between columns (adapters) are used to improve continuous learning. However, they report their results using the Asynchronous Advantage Actor-Critic (A3C) algorithm \cite{mnih2016asynchronous}. We could not find any evidence indicating whether they initialized separate columns and adapters for the value (critic) network as well as the policy (actor) network. Additionally, we could not locate any official implementation online, nor any implementation using actor-critic methods.

Intuitively, we expect that if PNN works for the policy, it should also work for the value function. Therefore, we implemented separate \textit{columns} for both the critic and actor networks, along with adapters for each new task. We encountered another issue: PPO is generally expected to outperform A3C in ProcGen environments. Thus, comparing HOP-PPO or base PPO with PNN-A3C would be unfair to PNN. To address this, we modified the PNN implementation to use a PPO update that propagates through separate columns and adapters.

Since the inputs for ProcGen environments are image-based, we use a single ReLU-activated convolutional layer as each adapter network. The input to the adapter network is the final convolutional output from the Critic or Actor of each column. The adapter outputs of the previous column are added to the original output and passed to the fully connected layers. All adapters are included in the gradient graphs to promote transfer, but the actor and critic columns are not included unless they are the active column.

[Upon publication, we will release our code for PNN-PPO].

\subsection{Hierarchical Weighting Examples}
\label{app:hw_examples}
\textbf{Hierarchical weightings.} This hierarchical structure implies that as the agent continues learning, the contributions of older policies diminish if more recent checkpoints are activated. Conversely, if more recent policies do not activate, the older policies will have a stronger influence. For example, consider the policy at the fourth checkpoint ($\pi_4$) as depicted in Figure \ref{fig:hierarchy_layout}. If activations $A_{s3}$, $A_{s2}$, and $A_{s1}$ occur, the policy output for the current state $s_t$ is given by:

\[
\pi_{4a}(s_t) = \frac{9}{24}\pi_{1_{\text{max}}}(s_{1_{\text{max-sim}}}) + \frac{1}{2}\pi_{2_{\text{max}}}(s_{2_{\text{max-sim}}}) + \frac{1}{2}\pi_{3_{\text{max}}}(s_{3_{\text{max-sim}}}) + \pi_4(s_t)
\]

In contrast, if only $A_{s1}$ activates, the output becomes:

\[
\pi_{4a}(s_t) = \frac{1}{2}\pi_{1_{\text{max}}}(s_{1_{\text{max-sim}}}) + \pi_4(s_t)
\]

Here, the contribution of $\pi_1$ diminishes as more recent policies are activated. However, if only $A_{s1}$ is activated, $\pi_1$ provides a significant contribution, which remains substantial regardless of how many other checkpoint policies exist but are inactive.

\end{document}